\documentclass[11pt,a4paper,hyphens]{article}

\author{Simon Hengchen,$^{\spadesuit}$\thanks{ ~ All authors contributed equally, and the ordering is determined in a round robin fashion.}  ~Nina Tahmasebi,$^{\spadesuit}$ Dominik Schlechtweg,$^{\clubsuit}$ Haim Dubossarsky$^{\heartsuit}$ \\
{\small $^{\spadesuit}$University of Gothenburg ~~ $^{\clubsuit}$University of Stuttgart ~~ $^{\heartsuit}$University of Cambridge}\\
{{\small \tt $^{\spadesuit}$firstname.lastname@gu.se ~~ $^{\clubsuit}$schlecdk@ims.uni-stuttgart.de ~~$^{\heartsuit}$hd423@cam.ac.uk}}\\
}

\title{Challenges for Computational Lexical Semantic Change}

\usepackage{tabularx}

\usepackage{enumitem} 
\usepackage{caption} 
\usepackage{subcaption} 
\usepackage[hidelinks]{hyperref}
\usepackage{xcolor}
\hypersetup{
    colorlinks,
    linkcolor={red!50!black},
    citecolor={blue!50!black},
    urlcolor={blue!80!black}
}

\usepackage{natbib}
\usepackage{graphicx}
\usepackage{tgpagella}
\usepackage{fancyhdr}
\usepackage{lastpage}

\begin{document}
\thispagestyle{fancy}

\maketitle

\abstract{
The computational study of lexical semantic change (LSC) has taken off in the past few years and we are seeing increasing interest in the field, %
from both computational sciences and linguistics.  
Most of the research so far has focused on methods for modelling and detecting semantic change using large diachronic textual data, with the majority of the approaches employing neural embeddings. While methods that offer easy modelling of diachronic text are one of the main reasons for the spiking interest in LSC, neural models leave many aspects of the problem unsolved. The field has several open and complex challenges. 
In this chapter, we aim to describe the most important of these challenges and outline future directions.
}

\section{Introduction}
\label{s:introduction}
The goal of tackling lexical semantic change (LSC) computationally is primarily to reconstruct semantic change evident in large diachronic corpora. 
The first papers addressing LSC appeared in 2008--2009 and since then a few papers per year have been published.\footnote{`Language evolution', `terminology evolution', `semantic change', `semantic shift' and 'semantic drift' are all terms that are or have been used for the concept which we denote \textit{lexical semantic change}.}
The first works that used neural embeddings were published in 2014 and 2015 \citep{kim2014temporal, kulkarni2015statistically, Dubossarsky2015} and together with \cite{Hamilton:2016}, they sparked interest in the research community for the problem of LSC.\footnote{Compare, for example, the roughly 30 papers at the start of 2018 as reported by \citet{tahmasebi2018survey}, with the roughly 50 papers submitted at the 1st International Workshop on Computational Approaches to Historical Language Change 2019 \citep{ws-2019-international-approaches}, and recent submissions at xACL venues, including the 21 papers submitted to the SemEval-2020 Task 1 on Unsupervised Lexical Semantic Change detection \citep{schlechtweg-etal-2020-semeval}.}  
Although there are a few research groups that have a longer  history of studying LSC using computational methods, the majority of  papers are single-entry papers where a group with an interesting model apply their method to a novel application on popular diachronic data. This leads to quick enhancement of methods but limits development and progress in other aspects of the field.

When surveying prior work, it is obvious that the computational field of LSC has been divided into two strands. 
The first strand deals with words as a whole and determines change on the basis of a word's dominant sense \citep[e.g.][]{kim2014temporal, kulkarni2015statistically}. 
An oft-used example is \texttt{gay}\footnote{More often than not, parts of speech are collapsed -- in this case, there is thus no difference between the adjective and the noun.} shifting from its `cheerful' sense to `homosexual.' 
The second strand deals with a word's senses\footnote{For the sake of clarity we use this as a simplified wording. We do not imply a fixed number of senses exist in a sense inventory; instead senses can overlap and be assigned different strengths.} individually -- for example, the `music' sense of \texttt{rock} has gradually come to describe not only music but also a certain lifestyle, while the `stone' sense remained unchanged (as seen in the works of \citealt{tahmasebi2013models} and \citealt{mitra2015automatic}).  
The first strand took off with the introduction of neural embeddings and its easy modelling of a word's semantic information. The second strand, faced with the immense complexity of explicitly modelling senses and meaning, has received much less attention. 

Computational models of meaning are at the core of LSC research, regardless of which strand is chosen. All current models, with the exception of those purely based on frequency, rely on the distributional hypothesis, which brings with it the set of challenges discussed in Section~\ref{s:theory}. But even accepting the distributional hypothesis and assuming meaning in context, the problem formulated by \cite{Schutze1998} remains: how does one accurately portray a word's senses? The question is valid regardless of whether the senses are represented individually or bundled up into one single representation. Recent developments in contextual embeddings \citep[e.g.][]{peters-etal-2018-deep} provide hope for accurate modelling of senses. However, they do not alleviate the problem of grouping sentence representations into sense correspondence. Within natural language processing (NLP), computational models of word meaning are often taken at face value and not questioned by researchers working on LSC. This is thus one of the areas that needs further attention in future work. Another area for thorough investigation is how useful sense-differentiation is for accurate LSC models. 

Another important area for future work is robust evaluation. Computational LSC methods model textual data as information signals and detect change in these signals. A signal can be a multidimensional vector, a cluster of words, topics, or frequency counts. 
This increasing level of abstraction is often ignored in evaluation; current evaluation standards allow for anecdotal evaluation of signal change, often without tying the results back to the text. Can we find evidence in the text for the detected changes?\footnote{To the best of our knowledge, only \citet{hengchen2017does} evaluates semantic change candidates output by a system reading a relatively large sample of sentences from the corpus studied -- but only for a single word, while several projects make use of extensive annotation to ensure that detected changes are present in the underlying textual corpus \citep{Lau12p591,schlechtweg-EtAl:2017:CoNLL,schlechtweg2018durel,schlechtweg-etal-2020-semeval,haettySurel:2019,perrone2019gasc,giulianelli-etal-2020-analysing}. }
So far, semantic annotation is the only way to evaluate methods on historical corpora while making sure that expected changes are present in the text. Annotating involves a significant investment of time and funds, and results in a limited test set.
A middle ground is to evaluate with respect to an outside source, like a dictionary or encyclopedia. However, while these resources offer an `expected' time and type of change, we can never be certain that these changes are reflected in the corpus under study. We refer to the example of \textit{computer} in \citep{tahmasebi2018survey}: the different parts of Google Books (British English, American English, and German) that reach the same level of frequency in different periods in time,  1934 for the German portion, 1943 for the American English, and 1953 for the British English. 
Recent work \citep{kulkarni2015statistically,dubossarsky2019timeout,shoemark2019room,SchlechtwegWalde20} introduced relatively cheap methods of generating \emph{synthetic} semantic change for any dataset, which we believe is an important path forward. The question is not if, but how synthetic evaluation data can complement costly manual evaluation. This will be answered in Section~\ref{s:evaluation}.

In addition, current methods work with rather coarse time granularity, largely because of the inherent complexity of adding multiple time bins (and senses) to the models. Unfortunately this constraint limits both the possibilities of the methods, and the results that can be found. Again, adding complexity to the models results in complexity in evaluation and calls for robust evaluation methods and data.  

In this chapter, we will discuss current and future challenges, and outline avenues for future work. More specifically, Section~\ref{s:data} discusses the requirements for textual resources, and their role for LSC. Section~\ref{s:theory} covers models of meaning, the current limitations of computational models, models of change, and what remains to be done in terms of time and sense complexity. 
Section~\ref{s:evaluation} sheds light on the need for robust evaluation practices and what we have learned so far. 
 We continue in Section~\ref{s:applications} to show the potential for collaboration between computational LSC and other fields, and end with some concluding remarks. 

\section{Data for Detecting LSC}
\label{s:data}
Hand in hand with the fast and simple modelling of word meaning, 
using neural embeddings for example, is the easy access to digital, diachronic texts that sparked mainstream interest in LSC as a problem domain for testing new models. For many reasons, including the early availability of large English corpora, there has long been a large over-representation of studies performed on English, in particular using COHA \citep{davies2002corpus}, Google N-grams \citep{michel2011quantitative}, and various Twitter corpora (see Table 2 in \citealt{tahmasebi2018survey} for an overview). As a consequence, most of the computational modelling of LSC has been developed and evaluated on these resources. However, tools and methods developed on one language (e.g., English) are not easily transferable to another language, a reoccurring challenge in other fields of NLP as well \citep{Bender:2011lilt, Ponti:2019cl}. Moreover, many languages may even lack the amount of historical, digitized data needed for robustly employing state-of-the-art methods like neural embeddings. This point is reinforced if we follow the recommendations of \cite{bowern2019semantic} for example, and distinguish groups based on features such as location, age, and social standing. The immediate result of this limitation is that many languages remain unstudied, or worse, studied with unsuitable methods. Consequently, whatever conclusions are drawn about LSC and formulated as `laws' are based on a very limited sample of languages, which may not be representative of the other 7,100 living languages.\footnote{Figure from ethnologue.com, \url{https://www.ethnologue.com/guides/how-many-languages}, last accessed 2020/01/16, rounded down to the lower hundred. In addition to living languages, dead languages are also studied, e.g. \citet{rodda2016panta-journal,perrone2019gasc,mcgillivray2019computational} for Ancient Greek.} 
As LSC mainly focuses on diachronic text, one obvious area for research is determining how well methods that rely on representations developed primarily for modern text transfer to historical languages.\footnote{See \citet{piotrowski2012natural} for a thorough overview of such challenges, and \citet{tahmasebi2013applicability} for an example of the applicability of a cluster-based representation on historical data.}\\

An important criterion for textual resources for LSC is good, reliable time stamps for each text. The texts should also be well distributed over longer time periods.\footnote{Recurring advertisements running for weeks at a time can effectively bias the corpus. See for example, the work by \citet[][pp.~67]{prescott2018searching} for a case study on the Burney newspaper collection.}  
For these reasons, newspaper corpora are popular for LSC studies. They also have the advantage that different news items are roughly equal in length.
But while they cover a large and interesting part of society, including technological inventions, they are limited in their coverage of everyday, non-newsworthy life.\footnote{Similar to news corpora, Twitter, another popular source for LSC research, offers posts which have timestamps and are consistent in size (though radically shorter than news articles), but with very different characteristics. However, most Twitter corpora are short-term, unlike the longer temporal dimensions of many news corpora.} 
On the other hand, large literary corpora like the Google N-grams pose different challenges including their skewed sampling of topics over time. For example, \citet{pechenick2015characterizing} point to an over-representation of scientific literature in the Google Books corpus, which biases the language used toward specific features mostly present in academic writing.
A second challenge to the use of the Google N-grams corpus is the small contexts (at most five words in a row) and the scrambled order in which these contexts are presented. To what extent this large resource can be used to study LSC remains to be investigated. One important question is whether changes found can be thoroughly evaluated using only these limited contexts.\footnote{Although there are several LSC papers using Google N-grams, e.g., \cite{wijaya2011understanding,Gulordava11}, to date there are no systematic investigations into the possibility of detecting different kinds of LSC, nor any systematic evaluation using grounding of found change in the Google N-grams.}

Other, less known corpora have other known deficits. For example, many literary works come in multiple editions with (minor) updates that modernize the language, while other texts lack known timestamps or publication dates. Some authors are more popular than others (or were simply more productive) and thus contribute in larger proportion and risk skewing the results.\footnote{See, for example, \cite{tangherlini2013trawling} where a single book completely changed the interpretation of a topic. Similarly, one can find many editions and reprints of the Bible in the Eighteenth Century Collections Online (ECCO) dataset that spans a century and contains over 180,000 titles; which will influence models. For a study on the effects of text duplication on semantic models, see \citet{schofield-etal-2017-quantifying}.}

What an optimal resource looks like is clearly dependent on the goal of the research, and LSC research is not homogeneous; different projects have different aims. While some aim to describe an interesting dataset (like the progression of one author), others want to use large-scale data to generalize to language outside of the corpus itself. In the latter case, it is important to be varied, as large textual corpora are not random samples of language as a whole (see, e.g., \citealt{Koplenig2016}) and whatever is encountered in corpora is only valid for those corpora and not for language in general. 

Most existing diachronic corpora  typically grow in volume over time. Sometimes this stems from the amount of data available. At other times it is an artefact related to ease of digitisation (e.g., only certain books can be fed into an automatic scanner) and OCR technology (OCR engines are trained on specific font families). This growth results in an extension of vocabulary size over time, which might not reflect reality and has serious effects on our methods. For example, previous work has shown that diachronic embeddings are very noisy \citep{hellrich2016bad, dubossarsky2017outta, dubossarsky2019timeout, kaiser-etal-2020-IMS, schlechtweg-etal-2020-semeval} with a large frequency bias, and are clearly affected by more and more data over time. 
Current and future studies are thus left with the question of whether the signal change we find really correspond to LSC in the text, or whether it is simply an artefact of the corpus.

We can only find what is available in our data: if we want to model other aspects of language and LSC, we need datasets that reflect those aspects well (for similar considerations related to evaluation, see Section~\ref{subsec:groundtruth}). This fact makes the case for using texts stemming from different sources, times, and places to allow for (re-)creating the complex pictures of semantic change. Thus general aspects beneficial for LSC are texts that are well-balanced (in time and across sources) and high-quality (with respect to OCR quality) with clear and fine-grained temporal metadata, as well as other kinds of metadata that can be of use. 
Until now, most existing computational LSC studies have been performed on textual data exclusively, aside from \citet{perrone2019gasc} and \citet{jawahar2019contextualized} who respectively used literary genre and social features as features. The reason for the under-utilisation of extra-linguistic metadata -- despite there being great need for it, as advocated by \citet{bowern2019semantic} -- is to a large extent the lack of proper and reliable metadata. In the case of Google N-grams, this kind of metadata is sacrificed in favour of releasing  large volumes of data freely. This path is also promising with respect to modelling the individual intent, described in Section~\ref{subsec:intent-indiv}. 

For the long-term future, we should raise the question of whether we  can model language at all using only texts \citep{bender2020climbing}. How much can we improve with multi-modal data in the future \citep{bruni2012distributional}, and what kind of data would be beneficial for LSC?

\section{Models of Meaning and Meaning Change}
\label{s:theory}

In this section, we shed light on the `meaning' we strive to model from the data, and on the challenges involved with modelling meaning computationally, and finally on how to employ the resulting information signals to establish change.

\subsection{Theory of lexical meaning and meaning change}
\label{sec:theorylex}

The field urgently needs definitions of the basic concepts it wants to distinguish: after all, we can draw from a rich tradition of semantics and semantic change research. The field traditionally starts with \citet{reisig1839vorlesungen}, although \citet{aristotle335BC} theorized metaphors in his Poetics well before then.\footnote{See for example the work by \citet{mague2005changements} for an overview.}

Here we focus on one theory which encompasses many others. \citet[p.~54][]{Blank97XVI} distinguishes three different levels of word meaning based on which type of knowledge a word can trigger in a human: (i) language-specific semantic, (ii) language-specific lexical, and (iii) language-external knowledge. The first comprises core semantic knowledge needed to distinguish different word meanings from each other.\footnote{Note that this level  covers only what \citeauthor{Blank97XVI} calls `knowledge' (p.~94). He then distinguishes six further levels of `meaning' (p.~94--96).} This knowledge corresponds to the minimal language-specific semantic attributes needed to structure a particular language, often called ``sememe'' in structural semantics. From these follow the hierarchical lexical relations between words (e.g. synonymy or hypernymy). The second level of word meaning comprises knowledge about the word's role in the lexicon (part of speech, word family or knowledge about polysemy/multiple meanings, referred to as `senses'' in this chapter). It includes the rules of its use (regional, social, stylistic or diachronic variety; syntagmatic knowledge such as selectional restrictions, phraseologisms or collocations). Level (iii) comprises knowledge about connotation and general knowledge of the world. 

\citeauthor{Blank97XVI} assumes that the knowledge from these three levels is stored in the mental lexicon of speakers, which can also change historically in these three levels (at least). An example of a change at the language-specific semantic level (i) is Latin \textit{pipio} `young bird' $>$ `young pigeon' which gained the attribute [pigeon-like] \citep[][p.~106--107]{Blank97XVI}. An example of change on the language-specific lexical level (ii) is \textit{gota} `cheek' which changes from being commonly used  in Old Italian to being used exclusively in the literary-poetic register in New Italian \citep[][p.~107]{Blank97XVI}. 
Finally, a change at the language-external knowledge level (iii) occurs when the knowledge about the referent changes. This can occur, for example, when the referent itself changes such as with German \textit{Schiff} `ship', as ships were primarily steamships in the 19th century, while today they are mainly motor ships \citep[][p.~111]{Blank97XVI}. 

Unfortunately, as will be made clear in the following subsection, this rich tradition of work is not used by the computational side of LSC because it is difficult to model meaning purely from written text. Currently, our modelling is very blunt. It can primarily capture contextual similarity between lexical items, and rarely distinguishes between different levels of meaning. Whether we draw from the large existing body of work that exists in traditional semantics research, or start from scratch with a new definition of what computational meaning is, we hope researchers in our field can come together and agree on what is, and should, be modelled. 

Similar to the conundrum 
in the definition of word meaning above, studies on LSC detection are seldom clear on the question of which type of information they aim to detect change in \citep{SchlechtwegWalde20}. 
There are various possible applications of LSC detection methods \citep[e.g.][]{Hamilton16,Voigt6521,kutuzov-etal-2017-temporal,hengchen2019nation}. Change at different levels of meaning may be important for different applications. For example, for literary studies it may be more relevant to detect changes of style, for social sciences the relevant level may be language-external knowledge and for historical linguistics the language-specific lexical and semantic levels may be more important.
Furthermore, LSC can be further divided into types (e.g., broadening/narrowing, amelioration/pejoration, metaphor and hyperbole). Several taxonomies of change have been suggested over the decades (\citealt{breal1897essai}, \citealt{Bloomfield1933} and \citealt{Blank199Whydonewmeaningoccur}, to name a few). Clearly, none of these applications or types of change can be properly tested until an adequate model of meaning is developed and the types of LSC to be investigated are meticulously defined.

\subsection{Computational models of meaning}

The need to choose the textual data available for the models and the decisions regarding the preprocessing  of the text are common to all models of computational meaning. While the influence of the former was described in Section~\ref{s:data}, and is fairly straightforward, extremely little attention is paid to preprocessing although its effects on the end results are far-reaching. 
The lower-casing of words often conflates parts of speech. For example \texttt{Apple} (proper noun) and \texttt{apple} (common noun) cannot be distinguished after lower-casing. Filtering out different parts of speech is also common practice, and can have radical effects on the results.\footnote{For discussions on the effects of preprocessing (or, as coined by \citealt{thompson-mimno-2018-authorless}, `purposeful data modification') for text mining purposes, we refer to \citet{schofield-mimno-2016-comparing}, \citet{schofield2017pulling}, \citet{denny2018text}, and \citet{tahmasebi2019strengths}.} Thus the effects of preprocessing on meaning representations should be investigated in the future.  

Nevertheless, the core of studying LSC computationally is the choice of the computational model of meaning: what we can model determines what change we can find. Crucially, methods for computational meaning inherit the theoretical limitations discussed in Section~\ref{sec:theorylex}. The challenge becomes even more cumbersome as existing methods for computational meaning rely on the distributional hypothesis \citep{harris1954distributional}, which represents word meaning based on the context in which words appear. In so doing, they often conflate lexical meaning with cultural and topical information available in the corpus used as a basis for the model. These limitations are not specific to semantic change, and lie at the basis of a heated debate that questions the fundamental capacity of computational models to capture meaning using only textual data, see for example \citet{bender2020climbing}.

There are different categories of computational models for meaning. These comprise a hierarchy with respect to the granularity of their sense/topic/concept representations:
 \begin{enumerate}[label=(\alph*)]
 \item a single representation for a word and all its semantic information (e.g., static embeddings),
 \item a representation that splits a word into semantic areas (roughly) approximating senses (e.g., topic models), and 
 \item a representation that models every occurrence of a word individually (e.g., contextual embeddings) and possibly groups them post-hoc into clusters of semantically-related uses expressing the same sense.
 \end{enumerate}

These categories of models differ with respect to their potential to address various LSC problems. For example, novel senses are hard to detect with models of category (a). However, these models have the upside of producing representations for all the words in the vocabulary, which is not the case for all models in category (b)  \citep[see for example][]{tahmasebi2013applicability}.  
In contrast, some sense-differentiated methods (category (b)), such as topic modelling allow for easy disambiguation so that we can deduce which word was used in which sense. However, category (a) models (e.g., word2vec) 
do not offer the same capability as they provide one vector per word, which is also biased toward the word's more frequent sense.\footnote{Every use of the word in a sentence is not accurately described by the vector representing it. In modern texts not pertaining to geology, a vector representation of \textit{rock} is biased toward its more frequent \textit{music} sense and will be a worse representation of a sentence where \textit{rock} is used in a \textit{stone} sense.}

Furthermore, models that derive meaning representations that can be interpreted and understood are needed to determine which senses of a word are represented, and whether they capture standard word meaning, topical use, pragmatics, or connotation (i.e., to distinguish between the levels of meaning referred to in Section~\ref{sec:theorylex}). The interpretability also allows us to qualitatively investigate different representations to determine which is better for different goals. 
 
Finally, the data requirements of our models can pose a critical limitation on our ability to model meaning (see Section~\ref{s:data}) as the computational models of meaning are data hungry and require extremely large amounts of text. They cannot be applied to the majority of the world's existing written languages as those often do not have sufficient amounts of written historical texts. If we follow the proposal of \cite{bowern2019semantic} and divide our data, not only by time, but also according to social aspects  \citep[to e.g. echo][]{meillet1905}, we reduce the amount of available data even further.
 
\subsection{Computational models of meaning change}

Change is defined and computed by comparing word representations between two or more time points, regardless of the specific model of meaning. 
Different models entail different mathematical functions to quantify the change. For example, and without claiming to be exhaustive: cosine or Euclidean distances are used for embedding models of continuous vectors representations, Hellinger distance and Jensen-Shannon or Kullback-Leibler divergences for topic distributions, and Jensen-Shannon divergence or cross-entropy for sense-differentiated representations. Ultimately the mathematical functions provide only a scalar that represents the degree of change. Determining change type from this information is not straightforward.
This impedes our ability to derive fine-grained information about the nature of change, as touched upon in Subsection~\ref{sec:theorylex}, and to incorporate theories of change which, for instance, postulate direction of change. 
For example, it becomes difficult to detect which sense changed, or to provide relevant distinctions related to the different applications (e.g., change in meaning vs. change in connotation), or taxonomies of change (e.g., broadening vs. narrowing). 
 \citet{SchlechtwegWalde20} identified two basic notions of change that are used in LSC research to evaluate the models' scalar change scores: (i) \textit{graded LSC}, where systems output to what degree words change \citep{Hamilton16,dubossarsky2017outta,bamler17,RosenfeldE18,rudolph2018dynamic,schlechtweg2018durel}, and (ii) \textit{binary LSC}, where systems make a decision on whether words have changed or not \citep{cook2014novel,tahmasebi2017finding,perrone2019gasc,shoemark2019room}. 
Despite this limitation of coarse scores of LSC, several change types have been targeted in previous research \citep[][see Table 4]{tahmasebi2018survey}. Importantly, in order for the results of LSC methods to be valuable for downstream tasks, we see a great need to determine the kind of change (e.g., broadening, narrowing, or novel sense). Methods that only detect one class, namely \textit{changed}, defer the problem to follow-up tasks: in which way has a word changed, or on what level (i)--(iii)\footnote{Though level (iii) relates to change in world-knowledge and goes well beyond semantic change.} from Section~\ref{sec:theorylex} did the change occur?

\subsubsection{Discriminating individual sentences}
\label{subsec:intent-indiv}
Meaning is ascribed to words at the sentence (utterance) level. However, for technical reasons related to the limitations of current computational models, previous work has carried out LSC only in large corpora. As a result, we model each word of interest with a signal (topic, cluster, vector) across all sentences and detect change in the signal. This discrepancy between the level at which the linguistic phenomena occur and the level of the analysis that is carried out may account for the type of questions commonly asked in contemporary research. In the majority of the cases, the signal change is evaluated on its own, and the question \textit{did the word meaning change or not?} is the only one answered. In a few rare cases, change is tied to the text and verified using the text. \textit{Did the word change in the underlying corpus or not?} is in fact a much more accurate question but is asked much less frequently. 
In a future scenario, where our models of computational meaning are much more fine-grained, we will be able to ask a third question: \textit{Is a specific usage of a word different than its previous uses?} To be able to tie the detected changes back to individual usage is much more demanding of any system and requires word sense discrimination (WSD) to be fully solved. Although radically more challenging, this task is also much more rewarding. It can help us in proper search scenarios, in dialogue and interaction studies, argument mining (where a person's understanding of a concept changes during the conversation), and in literary studies, to name but a few examples.

\subsubsection{Modelling of time}
The modelling of meaning change is directly dependent on the time dimension inherent in the data. Often, we artificially pool texts from adjacent years into long time bins because our computational models require large samples of text to produce accurate meaning representations or, to draw from research in historical sociolinguistics, because bins of a certain length are considered as `generations' of language users \citep{saily2016sociolinguistic}. Unfortunately, this leads to loss of fine-grained temporal information. 
From a modelling perspective, the inclusion of such information has the clear advantage of leading to more ecological models for LSC. This advantage can be used in two main ways: either to mitigate the noise associated with meaning representation models, or to detect regular patterns of change. Understandably, these advantages are only available when sufficient time points are included in the analysis. More time points, however, undoubtedly lead to greater computational complexity -- linearly if we consider the comparison of only subsequent time points, or quadratically if we consider all pairwise comparisons.

Some theories of LSC assume that change unfolds gradually through time, creating a trajectory of change  \citep[e.g., the regular patterns of semantic change in][]{traugott2001regularity}. Only models that acquire a meaning representation at several time points \citep[e.g.][]{tsakalidis-liakata-2020-sequential} are able to validate this underlying assumption by demonstrating a gradual trajectory of change. The work by \citet{RosenfeldE18} is an interesting example, as it models semantic change as a continuous variable and can also output the rate of change. Good extensions include allowing different change rates for different categories of words, or including background information about time periods where things change differently.  
In addition, models with multiple time points may contribute to improved LSC modelling by facilitating the discovery of intricate change patterns that would otherwise go unnoticed. 
For example, \citet{shoemark2019room} analysed Twitter data with high temporal resolution, and reported that several words demonstrated repeating seasonal patterns of change. The analysis of LSC trajectories easily lends itself to the use of modern change detection methods, which holds great promise for detecting hidden patterns of both change and regularities.

\section{Evaluation}
\label{s:evaluation}
Thus far, evaluation of LSC methods has predominantly ranged from a few anecdotally discussed examples to semi-large evaluation on (synthetic or pre-compiled) test sets, as made clear by Table 2 in \cite{tahmasebi2018survey}.\footnote{The work by \citet{hu-etal-2019-diachronic} uses \textit{dated} entries of the \textit{Oxford English Dictionary} and thus provides an exception.} The SemEval-2020 Task 1 on Unsupervised Lexical Semantic Change Detec\-tion provided the first larger-scale, openly available dataset with high-quality, hand-label\-ed judgements. It facilitated the first comparison of systems on established corpora, tasks, and gold-labels \citep{schlechtweg-etal-2020-semeval}. 

However, despite being the largest and broadest existing evaluation framework, the definition of LSC used in \citet{schlechtweg-etal-2020-semeval} -- i.e., a binary classification and a ranking task -- is a radical reduction of the full LSC task. The definition of LSC involves  modelling of words and detecting (sense) changes, as well as generalising across many more time points, and disambiguating instances of words in the text.
There cannot be only one universal model of a word: there are many ways to describe a word and its senses (see, for example, different dictionary definitions of the same word). So how do we devise evaluation data and methods such that different ways of defining meaning are taken into consideration when evaluating? Should a future evaluation dataset involve dividing the original sentences where a word is used in a particular sense into clusters with sentences that contributed to each sense, to avoid having to evaluate the different representations modelled for a word? How do we handle the uncertainty of which sense led to another? And how many new instances of change are needed to constitute semantic change? 

Current work in unsupervised LSC is primarily limited to binary decisions of `change' or `no change' for each word. However, some go beyond the binary to include graded change (although these changes are then often used in figures, for illustrative purposes), and a possible classification of change type. Future work in LSC needs to include a discussion of what role the modelling of sense and signal should play in the evaluation of semantic change: how large does the correspondence between the model and the `truth' for a model need to be, for the results to be deemed accurate? 
Should we be satisfied to see our methods performing well on follow-up (or downstream) tasks but failing to give proper semantic representation? 
Evaluation heavily depends on task definition -- and thus on the principle of fitness for use.\footnote{A concept originally from Joseph M. Juran, and thoroughly discussed in \citet{boydens1999informatique}.} In addition, to study LSC with different task definitions we need to have datasets that reflect these perspectives and make use of task-specific definitions of both meaning and change during evaluation.  

\subsection{Types of evaluations}
In the following subsections, we tackle two types of evaluation typically employed for LSC. We first discuss evaluation on ground-truth data, then tackle the promising evaluation on artificially-induced LSC data and argue that both should be used in a complementary fashion.

\subsubsection{Ground-truth}
\label{subsec:groundtruth}
An important part of evaluation is determining what to evaluate. 
For example, some studies perform quantitative evaluation of regularities in the vocabulary as a whole.
Regardless of other potential evaluation strategies, all existing work (also) evaluates change detected for a small number of lexical items -- typically words -- in a qualitative manner. 
This is done in one of two ways: either (i) a set of predetermined words are used for which there is an expected pattern of change, or (ii) the (ranked) output of the investigated method or methods is evaluated. Both of these evaluation strategies have the same aim, but with different (dis-)advantages, which we discuss below. 

(i) This evaluation strategy consists of creating a pre-chosen test set and has the advantage of requiring less effort as it removes the need to conduct a new evaluation for each change made to parameters such as size of time bins, or preprocessing procedure. The downside is, however, that the evaluation does not allow for new, previously unseen examples.\footnote{It is, of course, always possible to augment this `gold set' with new examples. Gold truth creation, though, is extremely costly both in time and money: \citet{schlechtweg-etal-2020-semeval} report a total cost of EUR 20,000 (1,000 hours) for 37 English words, 48 in German, 40 in Latin, and 31 in Swedish.} 
The pre-chosen words can be positive examples (words known to have changed), or negative examples (words known to be stable). Evaluation on only one class of words, positive or negative, does not properly measure the performance of a method. Let us say that we have a method that always predicts change, and we only evaluate on words that have changed. Unless we also evaluate exactly how the word has changed, or when, the method will always be 100\% accurate. The best indicator of a method's performance is its ability to separate between positive and negative examples, and hence any pre-chosen test set should consist of words from both classes. However, we also need a proper discussion of the proportion of positive and negative examples in the test set, as the most likely scenario in any given text is `no change'. 
	
(ii) Evaluating the output of the algorithm allows us to evaluate the performance of a method `in the wild' and truly study its behaviour. Unfortunately, this evaluation strategy requires new evaluation with each change either to the method or the data, as there potentially can be a completely new set of words to evaluate each time. 
The words to evaluate can be chosen on the basis of a predetermined measure of change (e.g., largest / smallest cosine angle between two consecutive time periods, i.e., the words that changed the most or least), or a set of randomly chosen words. Once a set of words is determined, the evaluation of each word is done in the same manner as for the pre-chosen test set. 
	
The \textit{accuracy} of the evaluation, regardless of strategy chosen, depends on the way we determine if and how a word has changed. The ground-truth must be constructed from the data (corpus) on which the methods are trained because existing dictionaries might list changes seen in the \emph{language}, that might not be present in the corpus, or vice versa. Requiring a method to find change that is not present in the underlying text, or considering detected changes as false because they are not present in a general-purpose dictionary, both lead to artificially low performance of the method. When (manually) creating ground-truth data for evaluation, sample sentences from the dataset should be read and taken into consideration, thus grounding the change in the dataset.

\subsubsection{Simulated LSC} 

Obtaining ground-truth data for LSC is a difficult task as it requires skilled annotators and takes time to produce. 
The problem is exacerbated as the time depth of the language change phenomena increases and the languages at hand become rarer. This fact leads to a further requirement: expert annotators. The notion of `expert annotator' is problematic when judging senses in the past. Previous studies \citep[e.g.][]{schlechtweg2018durel} note that historical linguists tend to have better inter-annotator agreements between themselves than with `untrained' native speakers -- hinting at the fact that this is a skill that can be honed. The difficulty of engaging sufficiently many expert annotators is also a theoretical argument in favour of synthetic evaluation frameworks as a complement.
In addition, some types of LSC are less frequent than others,\footnote{Assuming that semantic change is power-law distributed, like most linguistic phenomena. 
} 
therefore requiring large amounts of text to be annotated in order to find enough samples. To alleviate these problems, simulating LSC in existing corpora has been suggested.
 
 Simulating LSC is based on a decades-old procedure of inducing polysemy to evaluate word sense disambiguation systems \citep{Gale1992, Schutze1998}. 
 In this approach two or more words (e.g., \texttt{chair} and \texttt{sky}) are collapsed into a single word form (\texttt{chairsky}), thus conflating their meanings and creating a pseudo polysemous word (the original pair is removed from the lexicon). 
 From another perspective, if this procedure unfolds through time (i.e., a word either gained or lost senses), then it can be considered to simulate LSC via changes to the number of senses of words. 
 Indeed, this approach has been used extensively to simulate LSC \citep{Cook10, kulkarni2015statistically, RosenfeldE18, shoemark2019room}. 
 However, the LSC that is induced in this way is rather synthetic, because it collapses unrelated words into a single word form, as opposed to the general view that finds the different senses to be semantically related \citep{fillmore2000describing}. 
 In order to provide a more accurate LSC simulation, \citet{dubossarsky2019timeout} accounted for the similarity of candidate words prior to their collapse, and created both related pairs (e.g., \texttt{chair} and \texttt{stool}) that better reflect true polysemy, and compared the pair with the original approach of unrelated pairs (\texttt{chair} and \texttt{sky}). \citet{SchlechtwegWalde20} use SemCor, a sense-tagged corpus of English, to control for the specific senses each word has at each time point, thus providing an even more ecological model for simulated LSC.
 
 The simulated approach to LSC has the potential to circumvent any bottleneck related to the need for annotators, and thus reduces costs. 
 In addition, with careful planning, it should be possible to simulate any desirable type of LSC, regardless of its rarity in natural texts. As an added bonus, and certainly of interest to lexicographers, such an evaluation allows us to compute recall. In this scenario, recall would be proportional to the number of changed words that a given method can find. 
   Using a synthetic change dataset is currently  the only realistic scenario for determining the recall of our models and therefore, detecting how much change a method is able to capture. At the same time, it is hard to argue against the legitimate concern that these LSCs are artificial, and as such may not be the optimal way to evaluate detection by computational models. Certainly, synthetic change datasets are not optimal to study the natural linguistic phenomenon of semantic change, at least before we have a full understanding of the large-scale phenomena that we wish to study at which point we might no longer be in need for synthetic datasets.  
   However, without the considerable effort to annotate full datasets, we are bound to use synthetic change evaluation sets -- despite the inherent limitation described above. 
  As a result, an important factor for future research becomes the creation of synthetic datasets  that  reflect the complex and varying nature of real language and real semantic change. 
  
  We stipulate that simulated datasets should be used alongside ground-truth testing, both with respect to pre-chosen test sets, as well as evaluating the output, to properly evaluate the ability of any method to detect LSC.

\subsection{Quantifying meaning and meaning change}

 \begin{figure}[]
     \begin{subfigure}{0.495\textwidth}
 \frame{        \includegraphics[width=\linewidth]{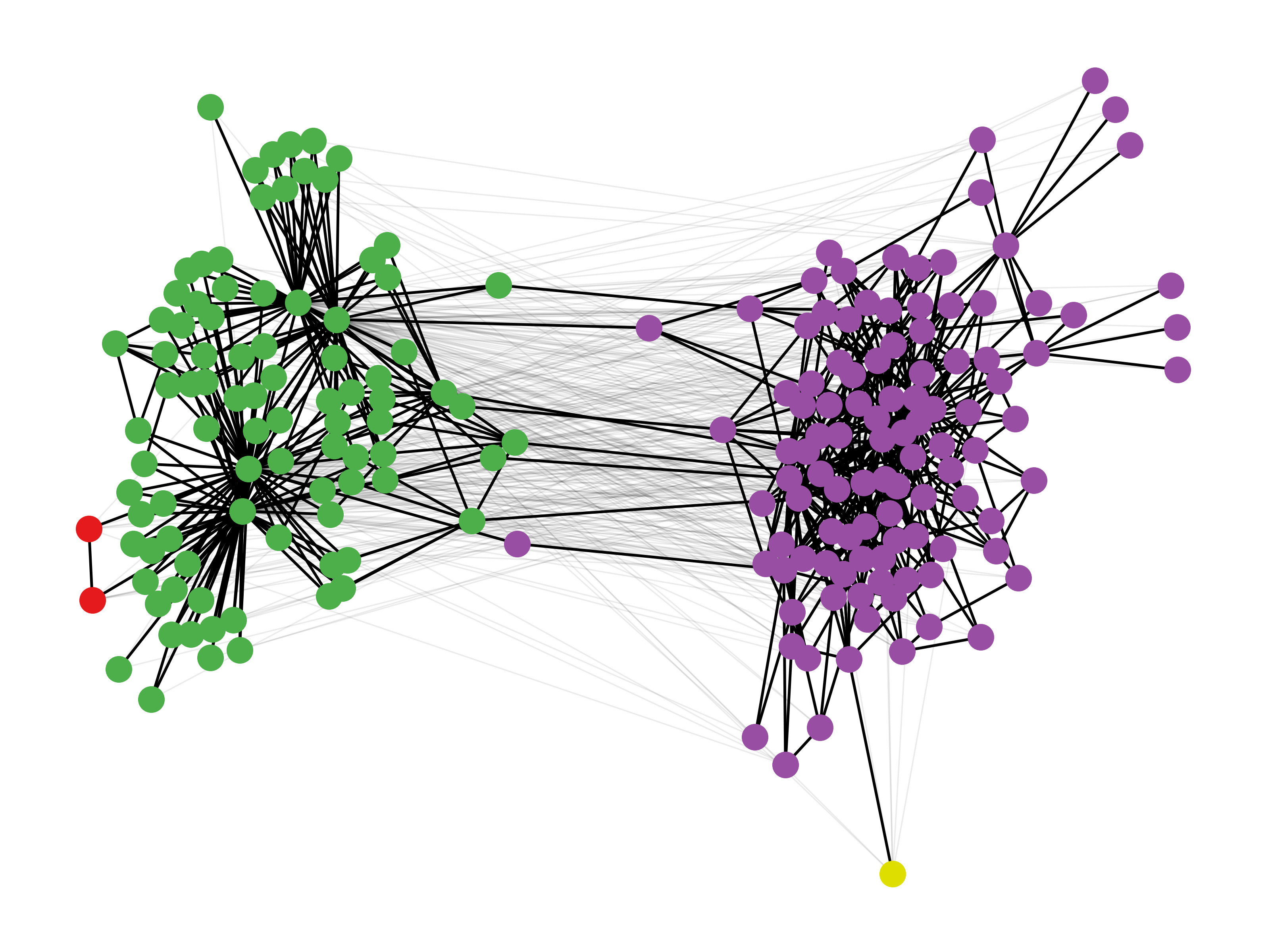}}
     \end{subfigure}
     \begin{subfigure}{0.495\textwidth}
 \frame {        \includegraphics[width=\linewidth]{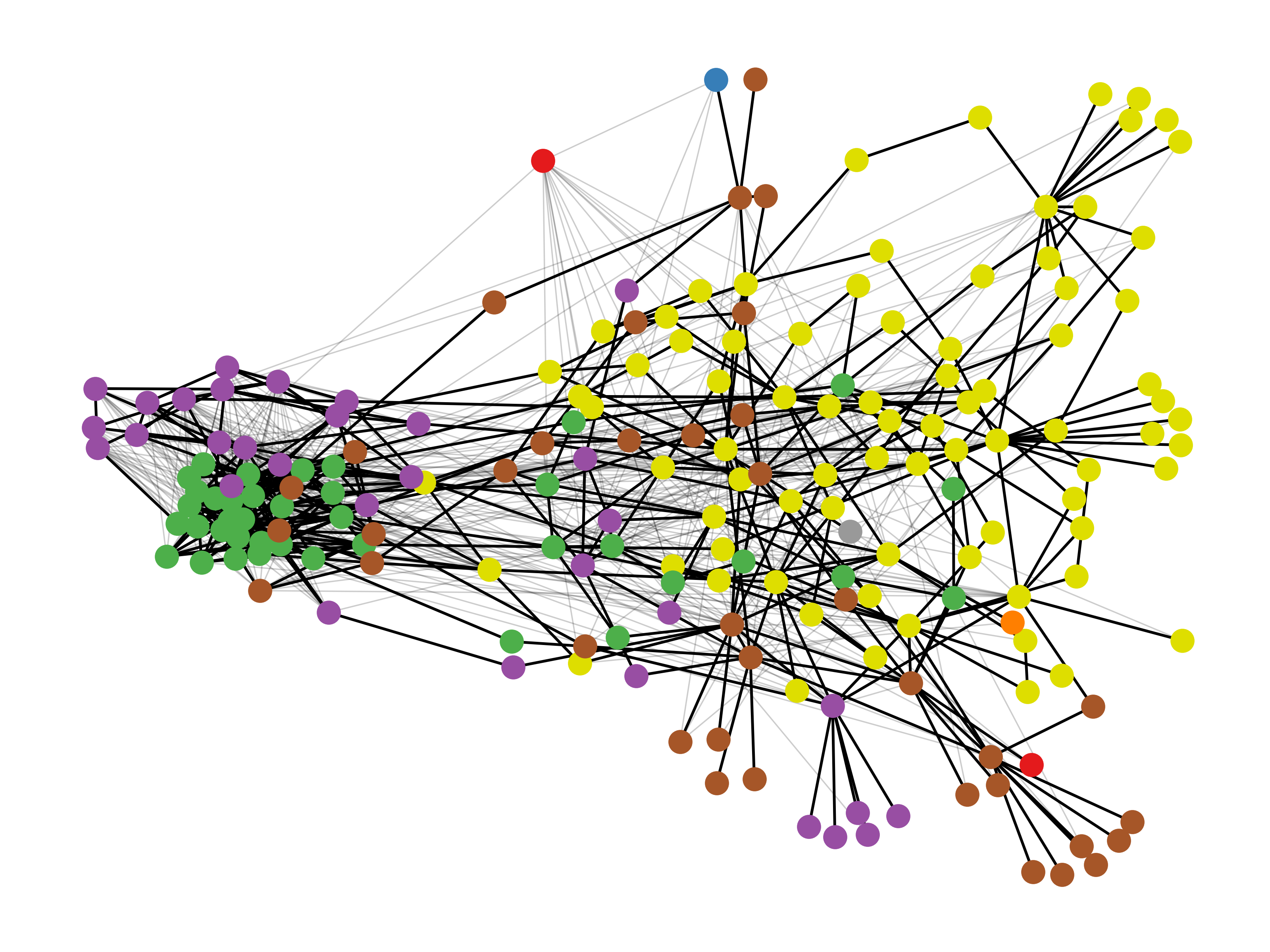}}
     \end{subfigure}
     \caption{Word Usage Graphs of German \textit{zersetzen} (left) and \textit{Abgesang} (right).}\label{fig:graph1}
\end{figure}

To provide high-quality, ground-truth data for LSC where word meaning and change is grounded in a given corpus, we must perform manual annotation. However, first, we need to choose the relevant level of meaning so that we can quantify meaning distinctions and the change for a word based on the annotation.
 Recently, SemEval-2020 - Task 1 \citep{schlechtweg-etal-2020-semeval} implemented a \textit{binary} and a \textit{graded} notion of LSC \citep{SchlechtwegWalde20} in the shared task, which was partly adopted by a follow-up task on Italian \citep{diacrita_evalita2020}. The annotation used typical meaning distinctions from historical linguistics \citep{Blank97XVI}. Although the authors avoided the use of discrete word senses in the annotation by using graded semantic relatedness judgements \citep{Erk13,schlechtweg2018durel}, they grouped word uses post-hoc into hard clusters and interpreted all uses in a cluster as having the same sense. 
 While this discrete view can work well in practice for some words \citep{OntoNotes2006}, it is inadequate for others \citep{Kilgarriff1997,carthy16}. In order to see this, consider Figure \ref{fig:graph1}, showing the annotated and clustered uses for two words from the SemEval dataset: the uses of the word \textit{zersetzen} on the left can clearly be partitioned into two main clusters, while the ones of \textit{Abgesang} on the right have a less clearly clusterable structure. 

 A graded notion of meaning and change can be used to avoid having to cluster cases like the latter, though it is still unclear what the practical applications could be for LSC without discrete senses. The advantage of discrete word senses is that, despite their inadequacy for certain words, they are a widely used concept and also build a bridge to historical linguistics \citep{Blank97XVI,blank1999historical}. 
 This bridge is an important one, because the most straightforward application of LSC detection methods is for historical linguistics or lexicography. Nonetheless, there might be many task-specific definitions of LSC that could do without sense distinctions, and the issue is an interesting avenue for future work. 

\section{Related fields and applications}
\label{s:applications}

\label{sec:related}

The field of LSC has close ties with two types of disciplines: those that study (i) \textbf{language}, and those that study (ii) \textbf{human activities}.
In this section, we shed light on prominent work in these fields without claiming to be exhaustive, and discuss the potential of interactions with these fields.

\subsection{Studying language}

A great deal of  existing work has gone into the study of language. 
Lexicography benefits a great deal from semantic representation in time, with works by, among others, \citet{Lau12p591},  \citet{Falk2014}, \citet{10.1093/ijl/ecy011}, \citet{klosaluengen}, and \citet{torresrivera2020detecting}.
In this strand, methods for LSC can prove efficient at updating historical dictionaries: by using LSC approaches on large-scale corpora, it becomes possible to verify, at the very least, whether a sense was actually used \emph{before} its current date in the dictionary. 
Senses cannot be post-dated, on the other hand; their absence from a corpus does not necessarily mean they did not exist elsewhere. 
Lexicographers can ideally use these methods to generate \emph{candidates} for semantic change which would then be manually checked. They could also use sense-frequency data to paint the prominence of a word's senses through time, or even incorporate a quantified measure of similarity between senses of the same word -- features that could also be incorporated in contemporary dictionaries.

Another strand, despite most work focusing solely on English, concerns language in general. 
In the past few years, there have been several attempts at testing hypotheses for laws of change which were proposed more than a century ago, or devising new ones.
\citet{Xu15} focus on two incompatible hypotheses: \citet{breal1897essai}'s law of differentiation (where near-synonyms are set to diverge across time) and \citet{stern1921swift}'s law of parallel change (where words sharing related meanings tend to move semantically in the same way). They showed quantitatively for English, in the Google Books corpus, that Stern's law of parallel change seems to be more rooted in evidence than Bréal's. 
\citet{Dubossarsky2015} ground their law of prototypicality on the hypothesis of \citet{geeraerts1997diachronic} that a word’s relation to the core prototypical meaning of its semantic category is crucial with respect to diachronic semantic change, and show using English data that prototypicality is negatively correlated with semantic change.
\citet{Eger:2016} postulate and show that semantic change tends to behave linearly in English, German and Latin.
Perhaps the best-known example of such work within NLP, and often the solely cited, are the two laws  of \citet{Hamilton:2016}: conformity (stating that frequency is negatively correlated with semantic change), and innovation (hypothesising that polysemy is positively correlated with semantic change).

Interestingly, since the NLP work above derives from observations that are replicable, quantitative, somewhat evidentiary, and not from a limited set of examples as was the case in the early non-computational days of semantic change research, previous laws elicited from quantitative investigations can be revisited. 
Such was the aim of \citet{dubossarsky2017outta}. They show that three previous laws (the law of prototypicality of \citealt{Dubossarsky2015}  and the laws of innovation and conformity by \citealt{Hamilton:2016}) are a byproduct of a confounding variable in the data, namely frequency, and are thus refuted. The paper calls for more stringent standards of proof when articulating new laws -- in other words, robust evaluation.

As regards future work, we envision the field of LSC moving towards better use of linguistic knowledge.
Traditional semantics and semantic change research is deeply rooted in theories that can now be computationally operationalized. 
Additionally, advances in computational typology and cross-lingual methods allow  language change to be modelled for several similar languages at the same time (as started by \citealt{uban-etal-2019-studying} and \citealt{frossard-etal-2020-dataset}, for example), and to take into account theories of language contact.
Other linguistic features can also be taken into account, and we hope to see more work going beyond `simple' lexical semantics.\footnote{An excellent example of this move forward can be seen in \citet{fonteyn2020grammar}, for example.}
The overview and discussions in this chapter have primarily targeted semantic change, often referred to as semasiological change in linguistics literature, while onomasiological change relates to different words used for the same concepts at different points in time. This general concept is often referred to as lexical replacement \citep{tahmasebi2018survey}. Future work should attempt to resolve onomasiological and semasiological change in an iterative manner to ensure coherency in our models. 


\subsection{Studying human society}

Along with the study of language itself, NLP techniques can be repurposed to serve different goals.
With NLP methods maturing and technical solutions being made available to virtually anyone,\footnote{For example, extremely large-scale pretrained models are shared on platforms such as HuggingFace's (\url{https://huggingface.co/models}) allowing anyone to download and use them with limited hardware; while efficient libraries such as gensim \citep{rehurek_lrec} make the training of type embeddings possible on personal laptops.} theories in other fields can be tested quantitatively. 
Quite obviously, since language is humans' best tool of communication, advanced techniques that tackle language are useful in many other fields, where they are often applied as-is, and sometimes modified to serve a different purpose.
What is often disregarded in NLP, however, is what we need from those tangential disciplines in order to arrive at reliable models.
One obvious answer to this question pertains to data resources. Those who are working on semantic change computation are heavily dependent on the data they receive, and should pay more attention to the type, diversity and quality of data they are working with, as discussed in Section~\ref{s:data}.

In this subsection we focus on a few examples of how related fields have borrowed methods from LSC by using some examples from the literature, and attempt to give broad avenues for a continued mutualistic relationship between LSC and those fields.

A great deal of past human knowledge that has survived is stored in texts.
Historical research\footnote{For clarity's sake, we do not differentiate between  `historical research', `digital history', `computational history', and `digital humanities.' For a broader discussion about field-naming in the (digital) humanities, refer to \cite{piotrowski_2020}.} is arguably a large benefactor of proper semantic representations of words in time: an often voiced critique in historical scholarship relates to chronological inconsistencies -- anachronisms \citep{syrjamaki_sins_2011}. 
As reported by \citet{zosa2020disappearing}, \cite{hobsbawm_history_2011} stated that `the most usual ideological abuse of history is based on anachronism rather than lies.' 
This fact leads to many historians trying to `see things their [the people of the past's] way' \citep{skinner_visions_2002}. 
Somewhat similarly, \cite{koselleck_vom_2010} underlines the `veto right of the sources.' However, for one to use the sources properly, they need to be understood correctly, and proper modelling of a word's semantics across time can definitely help historians interpret past events. 
Furthermore, the `concepts as factors and indicators of historical change' of \citet[p. 80]{koselleck2004futures}  highlights the importance of language as a window on the past.
There is a growing body of work with quantitative diachronic text mining (such as word embeddings and (dynamic) topic models) within humanities research which clearly benefits from NLP methods, but can similarly inform LSC. 
For example, \citet{heuser2017word}\footnote{See \url{https://twitter.com/quadrismegistus/status/846105045238112256} for a visualisation.} studies the difference between abstract and concrete words in different literary subgenres.
Similarly, Björck and co-authors\footnote{Presentation by Henrik Björck, Claes Ohlsson, and Leif Runefelt given at the Workshop on Automatic Detection of Language Change 2018 co-located with SLTC 2018, Stockholm. For more details, see \citet{ohlsson2020market}.} study the Swedish word for `market', \emph{marknad}, and describe a change in abstractness through time: from a physical market (as a noun), to more and more abstract notions (such as `labour market') and even to the point where the noun is used as an attributive (e.g. `market economy'). These observations teach us not only about the word itself, but also about the world.
If words such as `table' or `car' are relatively straightforward to define and probably easier to model (see e.g. \citealt{reilly2017effects} who show that concrete words tend to have denser semantic neighbourhoods than abstract words), what lessons can we learn from such work when representing abstract concepts?
LSC methods should strive to include such key information in its methods.

Claiming that current LSC methods can `solve historical research'\footnote{Or any field concerned with diachronic textual data.} and provide definitive answers to long-studied phenomena would be, at best, extremely misleading.
Indeed, while LSC methods can model a word's sense(s) across time, humanists (or political scientists, for that matter) can be described as studying \emph{concepts}. 
An emerging or evolving concept, almost by definition, will not be constrained to a single word. 
Rather, methods will probably have to be adapted to study a cluster of words\footnote{These clusters of words are related to what linguists call lexical fields, a term that in our experience is not widely used in other disciplines.} -- either manually chosen \citep{kenter2015ad,recchia2016tracing}, or selected in a more data-driven way \citep{tahmasebi2013models,hengchen2020vocab}. These clusters will be the basis for historical contextualisation and interpretation. 
The same ad-hoc adaptation is to be found in political science: a recent example of NLP methods making their way in (quantitative) political science is the work of \citet{rodman2020timely} where the author fits both an LDA model on more than a century of newspapers as well as a supervised topic model -- using 400 hand-annotated documents by several annotators with a high inter-annotator agreement -- so as to produce a gold standard to evaluate diachronic word embeddings, with the final aim of studying the evolution of \textit{concepts} such as `gender' and `race'. 
Similar work is undertaken by \citet{indukaev_studying_2021}, who studies modernisation in Russia and convincingly describes the benefits and limitations of topic models and word embeddings for such a study.

While extremely promising, our current methods fail to serve related fields that would benefit greatly from them: as of now, most LSC approaches simply model words, and not concepts -- again underlining the need for task-specific meaning tackled in Section~\ref{s:theory}. 

\section{Conclusions}
\label{s:conclusions}

In this chapter, we have outlined the existing challenges in reconstructing the semantic change evident in large diachronic corpora.

Currently, as was made obvious in Section~\ref{s:data}, the field suffers from several limitations when it comes to data. Indeed, we believe that future work should strive to use, and produce, high-quality text in many languages and different genres. 
This point is crucial: if we are to detect semantic change, our data needs to have certain precise qualities, as well as well-defined metadata. It is thus difficult for LSC researchers to rely on data created for other NLP fields.
As a plea to the larger community, we count on the field not to make the mistake of assuming that the available textual data is representative of the language at hand. 
We further hope that in the future, meaning can be modelled by using not only text, but also multi-modal data. 

Modelling is notoriously difficult, but, to paraphrase \citet{box1976science}, models being inherently wrong does not ineluctably make them useless.
A crucial component to the useful modelling of meaning and of change outlined in Section~\ref{s:theory} is the definition of what meaning is. 
Whether we draw from the large body of work that exists in traditional semantics research or start from scratch with a new definition of what \emph{computational} meaning is, we hope researchers in our field can come together and agree on \emph{what} is and should be modelled. Only with shared, solid models of meaning can the field move forward with the complexity, possibly intractable, of modelling meaning change. A word's semantics have changed -- but how?

Echoing the complexity of modelling information from data is the consistency needed in the evaluation of a model's output. Section~\ref{s:evaluation} makes the point that without a homogeneous, somewhat large-scale evaluation framework across languages such as the one proposed in \citet{schlechtweg-etal-2020-semeval}, researchers cannot confidently rely on conclusions from previous work to move forward.
Since ground-truth creation is expensive both in time and money and is ineluctably limited to a single corpus, we encourage the community to pay attention to synthetic evaluation techniques which have the potential to circumvent cost, evaluate different types of semantic change, and tackle different temporal granularities.
Our field is rich in methods but in dire need of comparable results. This can be partially solved with robust, thorough, and shared evaluation practices.

Being able to model and detect different types of semantic change is important in LSC, and also in related disciplines such as lexicography and historical linguistics. 
The history of ideas, and any area concerned with the diachronic study of textual data, would greatly benefit from our methods -- if they are robust. 
In addition, we believe that there is potential for a mutualistic relationship with those parallel fields not only contributing theory or domain expertise but also echoing the need for the proper modelling of words, senses, and types of change.

\section{Acknowledgements}
We thank our reviewers for their very helpful comments.
This chapter benefited from insightful comments from Lars Borin.
This work has been funded in part by the project \textit{Towards Computational Lexical Semantic Change Detection} supported by the Swedish Research Council (2019--2022; dnr 2018-01184), and \emph{Nationella Språkbanken} (the Swedish National Language Bank), jointly funded  by  the Swedish Research Council (2018--2024; dnr 2017-00626) and its ten partner institutions.
Dominik Schlechtweg was supported by the Konrad Adenauer Foundation and the CRETA Centre funded by the German Ministry for Education and Research (BMBF) during the writing of this chapter.
Haim Dubossarsky is supported by the Blavatnik Postdoctoral Fellowship Programme.
The authors wish to thank the Department of Swedish at the University of Gothenburg for providing financial support for language editing services.

\bibliography{localbibliography.bib}
\bibliographystyle{acl_natbib}

\end{document}